\pdfoutput=1

\documentclass[11pt]{article}

\usepackage{acl}

\usepackage{times}
\usepackage{latexsym}
\usepackage{booktabs}
\usepackage{graphicx}
\usepackage[T1]{fontenc}

\usepackage[utf8]{inputenc}

\usepackage{microtype}

\usepackage{color, colortbl}
\usepackage{amssymb}
\usepackage{balance}

\newcommand{\affA}{$^\diamondsuit$}
\newcommand{\affB}{$^\clubsuit$}

\title{Text Generation with Text-Editing Models}

\author{Eric Malmi\affA, Yue Dong\affB, Jonathan Mallinson\affA, Aleksandr Chuklin\affA, \\ \textbf{Jakub Adamek\affA, Daniil Mirylenka\affA, Felix Stahlberg\affA,} \\ \textbf{Sebastian Krause\affA, Shankar Kumar\affA, Aliaksei Severyn\affA} \\
  \affA Google \\
  \affB McGill University \& Mila \\
  \texttt{text-editing-tutorial@google.com} \\}

\begin{document}
\maketitle
\begin{abstract}
Text-editing models have recently become a prominent alternative to seq2seq models for monolingual text-generation tasks such as grammatical error correction, simplification, and style transfer. These tasks share a common trait -- they exhibit a large amount of textual overlap between the source and target texts. Text-editing models take advantage of this observation and learn to generate the output by predicting edit operations applied to the source sequence. In contrast, seq2seq models generate outputs word-by-word from scratch thus making them slow at inference time. Text-editing models provide several benefits over seq2seq models including faster inference speed, higher sample efficiency, and better control and interpretability of the outputs. This tutorial\footnote{Website: \url{https://text-editing.github.io/}} provides a comprehensive overview of text-editing models and current state-of-the-art approaches, and analyzes their pros and cons. We discuss challenges related to productionization and how these models can be used to mitigate hallucination and bias, both pressing challenges in the field of text generation.
\end{abstract}

\section{Introduction}

After revolutionizing the field of machine translation \citep{sutskever2014sequence,cho-etal-2014-learning,bahdanau2014neural}, sequence-to-sequence (seq2seq) methods have quickly become the standard approach for not only multilingual but also for \textit{monolingual} sequence transduction / text generation tasks, such as text summarization, style transfer, and grammatical error correction. While delivering significant quality gains, these models, however, are prone to hallucinations \citep{maynez-etal-2020-faithfulness,pagnoni-etal-2021-understanding}. The seq2seq task setup (where targets are generated from scratch word by word) overlooks the fact that in many monolingual tasks the source and target sequences have a considerable overlap, hence targets could be reconstructed from the source inputs by applying a set of edit operations.

\begin{figure}[tb]
\includegraphics[width=\columnwidth]{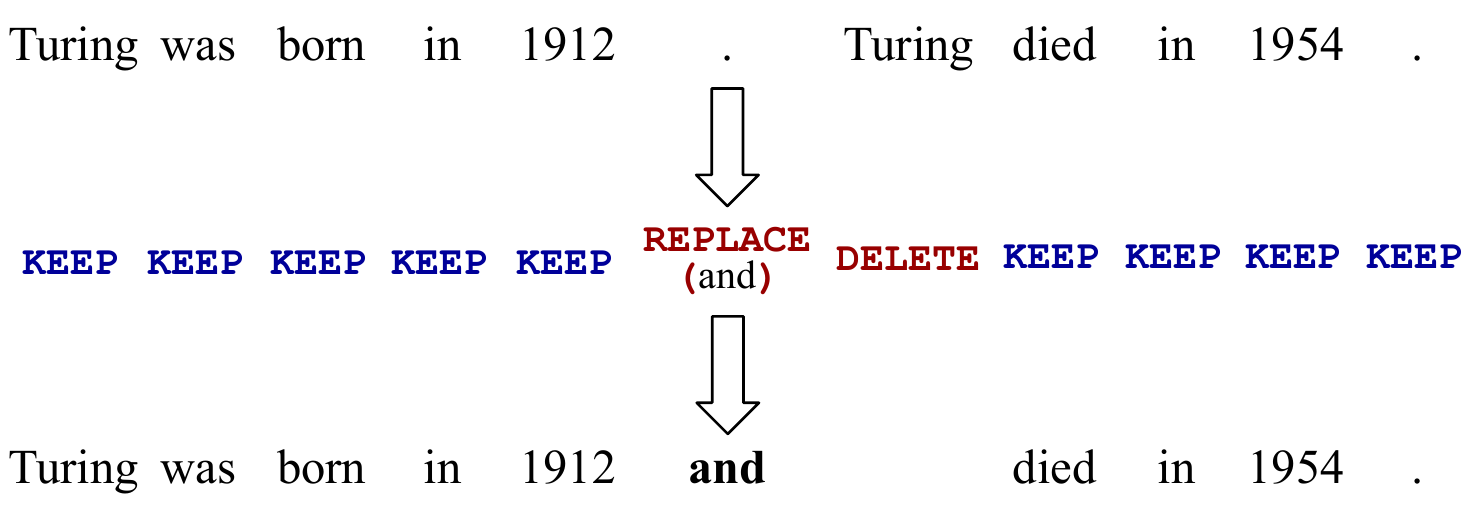}
\caption{An example of using a text-editing approach to solve a sentence-fusion task.}
\label{fig:example}
\end{figure}

Text-editing models attempt to address some of the limitations of seq2seq approaches and there has been recently a surge of interest in applying them to a variety of monolingual tasks including text simplification \citep{dong-etal-2019-editnts,mallinson-etal-2020-felix,agrawal-etal-2021-non}, grammatical error correction \citep{awasthi-etal-2019-parallel,omelianchuk-etal-2020-gector,malmi-etal-2019-encode,stahlberg-kumar-2020-seq2edits,rothe-etal-2021-simple,chen-etal-2020-improving-efficiency,hinson-etal-2020-heterogeneous,gao2021hierarchical}, sentence fusion \citep{malmi-etal-2019-encode,mallinson-etal-2020-felix} (see an example in Figure~\ref{fig:example}), 
MT automatic post-editing \citep{gu2019levenshtein,zietkiewicz2020post,mallinson-etal-2020-felix}, text style transfer~\cite{reid-zhong-2021-lewis,malmi-etal-2020-unsupervised}, data-to-text generation \citep{kasner-dusek-2020-data}, 
and utterance rewriting~\cite{liu-etal-2020-incomplete,DBLP:conf/sigir/VoskaridesLRKR20, Hct2022}.

Text-editing approaches claim to be more accurate or on-par with seq2seq baselines especially in low resource settings, less prone to hallucinations and faster at inference time.
These advantages have generated a substantial and continued level of interest in text-editing research. 
The goal of this tutorial is to provide the first comprehensive overview of the family of text-editing approaches and to offer practical guidelines for applying them to a variety of text-generation tasks.

\subsection{Target Audience and Prerequisites}

The tutorial is intended for researchers and practitioners who are familiar with generic seq2seq text-generation methods, such as Transformer~\citep{vaswani2017attention} and pre-trained language models like BERT~\citep{devlin-etal-2019-bert}. However, prior experience with text-editing models is not required to be able to follow the tutorial.

We expect the topic to attract people in both academia and industry. The high-sample efficiency and low-computational requirements of text-editing models~\citep{malmi-etal-2019-encode,mallinson-etal-2020-felix} makes them an attractive baseline, e.g., for researchers developing new text-generation tasks for which large training sets do not yet exist. Moreover, the high-inference speed of text-editing methods, owing to their often non-autoregressive architecture~\citep{awasthi-etal-2019-parallel,mallinson-etal-2020-felix}, makes them suitable for building real-time applications.

\section{Tutorial Outline}

The structure of the tutorial with duration estimates for different sections are shown in Table~\ref{tab:structure}. Below we provide brief descriptions for each section.

\begin{table}
\centering
\small
\begin{tabular}{lr}
\toprule
\textbf{Section} & \textbf{Duration} \\
\midrule
Introduction & 15 min \\
$\quad$ What are text-editing models? &  \\
$\quad$ Text-editing vs. seq2seq models &  \\
Model design & 40 min \\
$\quad$ Example model + model landscape &  \\
$\quad$ Edit-operation types &  \\
$\quad$ Tagging architecture &  \\
$\quad$ Auto-regressiveness &  \\
$\quad$ Converting target texts to target edits &  \\
Applications & 45 min \\
$\quad$ Overview &  \\
$\quad$ Grammatical Error Correction &  \\
$\quad$ Text Simplification &  \\
$\quad$ Unsupervised Style Transfer &  \\
$\quad$ Incomplete Utterance Rewriting &  \\
Controllable generation & 25 min \\
$\quad$ Mitigating hallucinations &  \\
$\quad$ Controllable dataset generation &  \\
Multilingual text editing & 25 min \\
$\quad$ Tokenization &  \\
$\quad$ Handling morphology &  \\
$\quad$ Practical aspects &  \\
Productionization & 25 min \\
$\quad$ Latency &  \\
$\quad$ Sample efficiency &  \\
Recommendations and future directions & 5 min \\
\midrule
\textbf{Total} & 180 min \\
\bottomrule
\end{tabular}
\caption{Tutorial structure and duration of each section.}
\label{tab:structure}
\end{table}

\paragraph*{Introduction.}

We first define the family of text-editing methods: Text-editing models are sequence-transduction methods that produce the output text by predicting edit operations which are applied to the inputs. In contrast, the traditional seq2seq methods produce the output from scratch, token by token. We summarize the main pros and cons of these two approaches and provide guidelines for choosing which approach is more suitable for a given task.

\paragraph*{Model Design.}

The similarities and differences of a set of popular text-editing methods will be analyzed in terms of the types of edit operations they employ, their tagging architecture, and whether they are auto-regressive or feedforward. We also discuss methods for converting target texts into target edit sequences, a task which often does not have a unique solution.
Table~\ref{tab:methods} provides a summary of the similarities and differences between the methods covered in the tutorial.

\begin{table*}
\centering
\small
\resizebox{\textwidth}{!}{%
\begin{tabular}{lcccccc}
\toprule
Method & \begin{tabular}[x]{@{}c@{}}Non-autore-\\gressive\end{tabular} & \begin{tabular}[x]{@{}c@{}}Pre-trained\\decoder\end{tabular} & \begin{tabular}[x]{@{}c@{}}Reorde-\\ring\end{tabular} & \begin{tabular}[x]{@{}c@{}}Unsuper-\\vised\end{tabular} & \begin{tabular}[x]{@{}c@{}}Language-\\agnostic\end{tabular} & Application(s) \\
\midrule
EdiT5 \citep{edit5} & (\checkmark) & \checkmark & \checkmark &  & \checkmark & \emph{multiple}
\\\rowcolor{gray!10}
EditNTS \citep{dong-etal-2019-editnts} &  &  &  &  & \checkmark & Simplification
\\
Felix \citep{mallinson-etal-2020-felix} & \checkmark & \checkmark & \checkmark &  & \checkmark & \emph{multiple}
\\\rowcolor{gray!10}
GECToR \citep{omelianchuk-etal-2020-gector} & \checkmark & (\checkmark) &  &  &  & GEC
\\
HCT \citep{Hct2022} & \checkmark & & \checkmark & & \checkmark & Utterance Rewriting
\\\rowcolor{gray!10}
LaserTagger \citep{malmi-etal-2019-encode} & \checkmark &  &  &  & \checkmark & \emph{multiple}
\\
LevT \citep{gu2019levenshtein} & (\checkmark) & \checkmark &  &  & \checkmark & \emph{multiple}
\\\rowcolor{gray!10}
LEWIS \citep{reid-zhong-2021-lewis} &  & \checkmark &  & \checkmark & \checkmark & Style Transfer
\\
Masker \citep{malmi-etal-2020-unsupervised} & \checkmark & \checkmark &  & \checkmark & \checkmark & \emph{multiple}
\\\rowcolor{gray!10}
PIE \citep{awasthi-etal-2019-parallel} & \checkmark & \checkmark &  &  &  & GEC
\\
Seq2Edits \citep{stahlberg-kumar-2020-seq2edits} &  &  &  &  & (\checkmark) & \emph{multiple}
\\\rowcolor{gray!10}
SL \citep{alva-manchego-etal-2017-learning} & \checkmark &  & \checkmark &  & \checkmark & Simplification
\\
\bottomrule
\end{tabular}
}
\caption{Overview of selected text-editing methods.}
\label{tab:methods}
\end{table*}

\paragraph*{Applications.}

A key criterion for determining whether text-editing models are a good fit for a given application is the average degree of overlap between source and target texts. The higher the overlap, the more input tokens can be reused to generate the target, thus resulting in a simpler edit sequence. We give an overview of applications with a high degree of overlap to which text-editing methods have been applied to. Then we do a deep dive in to the following applications: grammatical error correction, text simplification, unsupervised style transfer, and incomplete utterance rewriting.

\paragraph*{Controllable Generation.}

Text-editing models with a restricted vocabulary of phrases to insert \citep{malmi-etal-2019-encode, Hct2022} or with linguistically informed suffix-transformation operations \citep{awasthi-etal-2019-parallel,omelianchuk-etal-2020-gector} are less prone to different types of hallucination since the models cannot produce arbitrary outputs. Moreover, the restricted vocabulary makes it feasible to manually refine the list of phrases that the model can insert.
Another route through which the decomposition of the generation task into explicit edit operations can improve controllability is via biasing of certain types of edits to control how often the model will insert new text \citep{dong-etal-2019-editnts,omelianchuk-etal-2020-gector}.
Controllable generation with editing models can be useful for generating large synthetic datasets with a desired distribution of errors, which yields improvements in tasks such as grammatical error correction \citep{stahlberg-kumar-2021-synthetic}. We will provide concrete examples of the aforementioned control measures and their effects. 

\paragraph*{Multilingual Text Editing.}

Most text-editing models, like text-generation models in general, are evaluated on English, but there are also methods evaluated or specifically developed for other languages, including Chinese~\citep{hinson-etal-2020-heterogeneous,liu-etal-2020-incomplete}, Czech~\citep{naplava-straka-2019-grammatical},  German~\cite{mallinson-etal-2020-felix}, Russian~\citep{stahlberg-kumar-2020-seq2edits}, and Ukrainian~\citep{syvokon2021uagec}. 
Apart from general tokenization-related challenges discussed in~\cite{mielke2021between},
an additional challenge with applying text-editing methods to morphologically rich languages is a potential mismatch between the subword tokens, on which the underlying sequence labeling model operates, and the morphemes or affixes, on which the edits should happen. Possible solutions to this challenge include developing custom inflection operations~\citep{awasthi-etal-2019-parallel,omelianchuk-etal-2020-gector}
or learning them from the data~\cite{straka2021character},
and using more fine-grained edit operations, such as character-level edits~\cite{gao2021hierarchical}.

An additional challenge when building a truly multilingual model---as opposed to one model per language---is to ensure that it is not skewed towards a particular language or a set of languages~\citep{chung-etal-2020-improving} while being computationally efficient.

\paragraph*{Productionization.}
We discuss how casting a text-generation problem as a text-editing task often allows the use of significantly faster and more data-efficient model architectures, without sacrificing output quality. We make use of the TensorFlow Profiler\footnote{\url{https://www.tensorflow.org/guide/profiler\#trace_viewer_interface}} to compare latencies of text-editing and non-text-editing solutions for an example problem, and illustrate where the time savings come from.

\paragraph*{Recommendations and Future Directions.}

\begin{figure}[tb]
\includegraphics[width=\columnwidth]{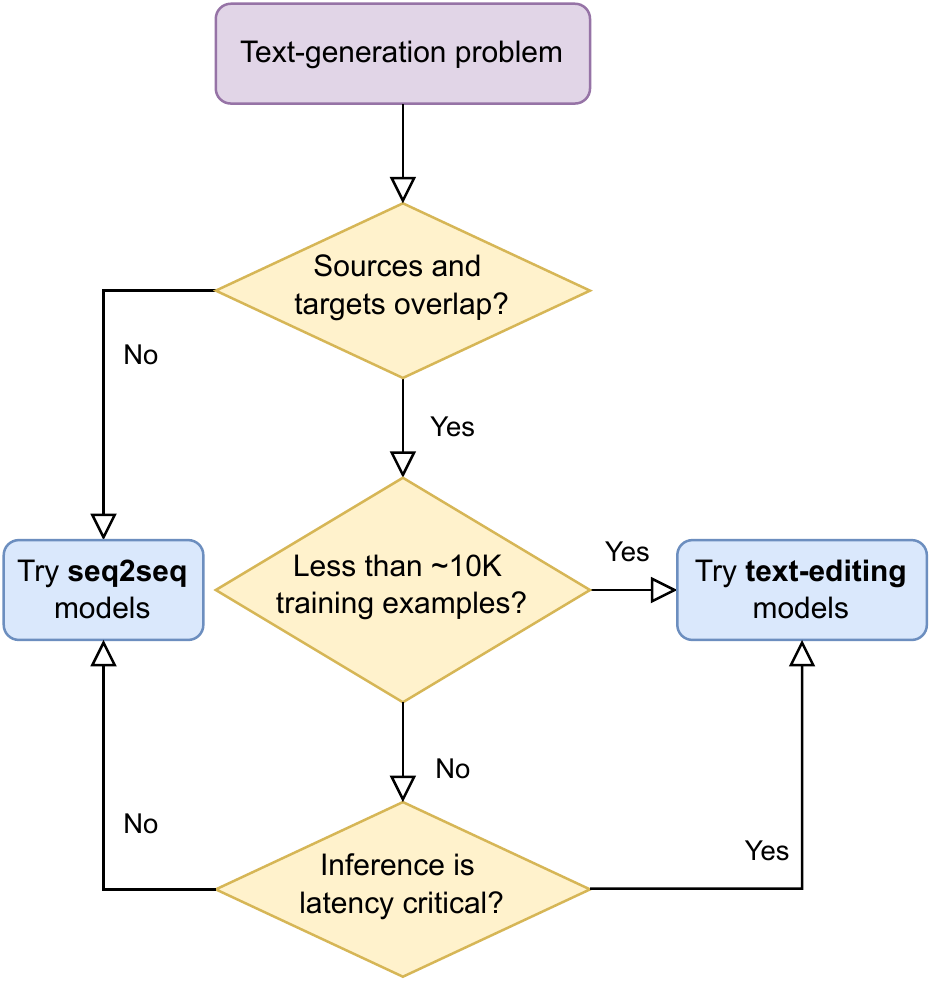}
\caption{Proposed flowchart for deciding when to try a text-editing approach.}
\label{fig:rec}
\end{figure}

We provide practical guidelines for when to use (and when not to use) text-editing methods (see Figure~\ref{fig:rec} for a summary). We also outline possible future directions which include: ($i$)~learned edit operations, ($ii$)~studying the effects of different subword segmentation methods since these typically determine the granularity at which the edit operations are applied, ($iii$)~text-editing-specific pre-training methods, ($iv$)~sampling strategies for text-editing methods, and ($v$)~studying the effects of scaling up text-editing methods, a strategy that has been found to be very effective for many other text-generation methods~\cite{brown2020language,chowdhery2022palm}.

\section{Diversity Considerations}

A significant portion of the tutorial is devoted to discussing multilingual text-editing, including applying text-editing models to morphologically rich languages which presents specific challenges related to larger vocabularies and the need to edit word affixes. The presenters come from both academia and industry, are native speakers of 8 languages based in 4 different countries (Switzerland, Germany, Canada, USA), and are of different seniority levels from a PhD student to a Senior Staff Research Scientist.

\section{Reading List}

Before the tutorial, we expect the audience to read \citep{vaswani2017attention} and \citep{devlin-etal-2019-bert}. For references to text-editing works that will be discussed in the tutorial, see Table~\ref{tab:methods}.

\paragraph*{Breadth.} 50\% of the methods that will be discussed in the tutorial (cf. Table~\ref{tab:methods}) are developed by different subsets of the tutorial instructors.

\section{Presenters}

\paragraph*{Eric Malmi} is a Senior Research Scientist at Google Switzerland. His research is focused on developing text-generation models for grammatical error correction and text style transfer. He received his PhD from Aalto University, Finland, where he also taught a course on Recent Advances in Natural Language Generation in Spring 2022.

\paragraph*{Yue Dong} 
is a final-year PhD student in CS at McGill University and Mila, Canada. Her research is focused on conditional text generation. She is a co-organizer for the NewSum workshop at EMNLP 2021 and ENLSP workshop at NeurIPS 2021.

\paragraph*{Jonathan Mallinson}  is a Research Engineer at Google Switzerland. His research is focused on low-latency text-to-text generation. He received his PhD from the University of Edinburgh, Scotland.

\paragraph*{Aleksandr Chuklin}  is a Research Engineer at Google Switzerland. His current research focuses on multi-lingual NLG. He organized \href{http://scai-workshop.github.io}{workshops} and conducted \href{https://clickmodels.weebly.com/tutorials.html}{tutorials} at conferences such as SIGIR, EMNLP, and IJCAI. Aleksandr received his PhD from University of Amsterdam, The Netherlands.

\paragraph*{Jakub Adamek}  is a Research Engineer at Google Switzerland focusing on grammatical error correction and low-latency models. He received his MSc from Jagiellonian University.

\paragraph*{Daniil Mirylenka} is a Research Engineer at Google Switzerland working on text editing with application to grammatical error correction. He received his PhD from the University of Trento, Italy.

\paragraph*{Felix Stahlberg}  is a Research Scientist at Google focusing on grammatical error correction and text style models. He received his PhD from Cambridge University, UK.

\paragraph*{Sebastian Krause} is a Senior Research Engineer at Google Switzerland. His work is focused on multi-lingual rewriting of questions in low-latency settings. Sebastian received his PhD in Engineering from the Technical University of Berlin, Germany.

\paragraph*{Shankar Kumar} is a Senior Staff Research Scientist at Google leading a research team working on speech and language algorithms. He received his PhD from the Johns Hopkins University, US.

\paragraph*{Aliaksei Severyn} is a Staff Research Scientist at Google Switzerland leading an applied research team working on next generation NLG solutions. He received his PhD from University of Trento, Italy.

\section{Ethical Considerations}

Text-generation methods have the potential to generate non-factual~\cite{maynez-etal-2020-faithfulness,pagnoni-etal-2021-understanding,kreps2020all} and offensive content~\cite{gehman-etal-2020-realtoxicityprompts}. Furthermore, training these models on uncurated data can lead to the models replicating harmful views presented in the training data~\citep{bender2021dangers}. Text-editing models are also susceptible to these issues, but they have been shown to mitigate some of them. Specifically, they reduce the likelihood of different types of hallucination~\citep{malmi-etal-2019-encode} and their higher sample efficiency~\cite{malmi-etal-2019-encode,mallinson-etal-2020-felix} enables more careful curation of the training data. The tutorial will discuss the ethical issues related to text generation and provide concrete examples on how text-editing models can help mitigate them.

\bibliography{anthology,custom}

\balance

\end{document}